\documentclass[10pt,twocolumn,letterpaper]{article}

\usepackage{cvpr}              
\usepackage{url}
\usepackage{algorithm}
\usepackage{algorithmic}

\usepackage{multirow}
\usepackage[table]{xcolor}
\usepackage{color}
\definecolor{mygreen}{RGB}{146, 199, 113} 
\definecolor{mypurple}{RGB}{146, 199, 113}
\usepackage{graphicx}
\usepackage{adjustbox}
\usepackage{amsmath}
\usepackage{array}
\usepackage{booktabs}
\usepackage{amssymb}
\usepackage{dsfont}
\usepackage{graphicx}
\usepackage{footmisc}
\usepackage{makecell}
\usepackage[accsupp]{axessibility}

\definecolor{cvprblue}{rgb}{0.21,0.49,0.74}
\usepackage[pagebackref,breaklinks,colorlinks,allcolors=cvprblue]{hyperref}

\def\paperID{3076} 
\def\confName{CVPR}
\def\confYear{2025}

\title{
    \centering
    DivControl: Knowledge Diversion for Controllable Image Generation
}

\author{Yucheng Xie\textsuperscript{\rm 1,2}\quad Fu Feng\textsuperscript{\rm 1,2}\quad Ruixiao Shi\textsuperscript{\rm 1,2}\quad Jing Wang\textsuperscript{\rm 1,2}\thanks{Corresponding authors}\quad Yong Rui\textsuperscript{\rm 3}\quad Xin Geng\textsuperscript{\rm 1,2}\footnotemark[1]\\
\textsuperscript{\rm 1}School of Computer Science and Engineering, Southeast University, Nanjing, China\\
\textsuperscript{\rm 2}Key Laboratory of New Generation Artificial Intelligence Technology and Its Interdisciplinary \\Applications (Southeast University), Ministry of Education, China\\
\textsuperscript{\rm 3}Lenovo Research\\
{\tt\small \{xieyc, fufeng, wangjing91, xgeng\}@seu.edu.cn}\\
}

\begin{document}
\maketitle

\begin{abstract}
    Diffusion models have advanced from text-to-image (T2I) to image-to-image (I2I) generation by incorporating structured inputs such as depth maps, enabling fine-grained spatial control.
    However, existing methods either train separate models for each condition or rely on unified architectures with entangled representations, resulting in poor generalization and high adaptation costs for novel conditions.
    To this end, we propose \textbf{DivControl}, a decomposable pretraining framework for unified controllable generation and efficient adaptation.
    DivControl factorizes ControlNet via SVD into basic components—pairs of singular vectors—which are disentangled  into condition-agnostic learngenes and condition-specific tailors through knowledge diversion during multi-condition training.
    Knowledge diversion is implemented via a dynamic gate that performs soft routing over tailors based on the semantics of condition instructions, enabling zero-shot generalization and parameter-efficient adaptation to novel conditions.
    To further improve condition fidelity and training efficiency, we introduce a representation alignment loss that aligns condition embeddings with early diffusion features.
    Extensive experiments demonstrate that DivControl achieves state-of-the-art controllability with 36.4$\times$ less training cost, while simultaneously improving average performance on basic conditions. 
    It also delivers strong zero-shot and few-shot performance on unseen conditions, demonstrating superior scalability, modularity, and transferability.
\end{abstract}

\section{Introduction}
Diffusion models have demonstrated remarkable performance in text-to-image (T2I) generation, with models such as DALL-E 3~\cite{ramesh2022hierarchical}, Stable Diffusion 3~\cite{esser2024scaling}, and Midjourney~\cite{midjourney} generating images comparable to human-created artwork from natural language prompts~\cite{balaji2022ediff, peebles2023scalable, zhang2023adding}. 
Their ability to produce high-fidelity, semantically aligned outputs has positioned diffusion models as a foundation for controllable generation. 
To support finer-grained and more deterministic control, recent efforts have extended conditions beyond text to structured visual inputs—such as depth and segmentation maps—enabling image-to-image (I2I) generation with explicit spatial guidance~\cite{zhang2023adding, zavadski2024controlnet, mou2024t2i}.

\begin{figure}[tb]
  \centering
  \includegraphics[width=\linewidth]{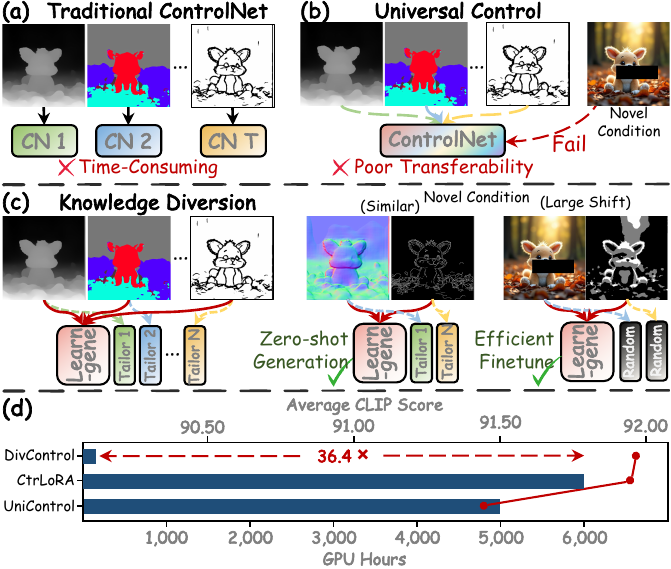}
  \vspace{-0.25in}
  \caption{(a)~Traditional ControlNet requires training a dedicated model for each control type, leading to substantial computational overhead. (b)~Universal control approaches aim to unify all conditions within a single model, but exhibit poor generalization to unseen tasks. (c)~DivControl addresses this by introducing knowledge diversion to disentangle condition-agnostic and condition-specific representations during training, enabling unified control and zero-shot transfer. (d)~This modular design reduces training cost by over \textbf{36.4$\times$} (165 vs. 6000 GPU hours) while achieving superior controllable generation quality.
  }
  \label{fig:intro}
  \vspace{-0.1in}
\end{figure}

However, training separate diffusion models for each control condition is computationally prohibitive.
For instance, ControlNet~\cite{zhang2023adding} requires over 600 A100 GPU hours on 3 million images to support a single \textsc{Canny} condition, with other modalities demanding even more resources~\cite{xu2024ctrlora}.
To improve efficiency and scalability, recent works propose unified frameworks for multi-condition control~\cite{qin2023unicontrol, tan2024ominicontrol, wang2025unicombine}, aiming to handle diverse conditions within a single model (Figure~\ref{fig:intro}a, b).
However, these approaches struggle to generalize to unseen or heterogeneous conditions, as jointly learned representations tend to be entangled, hindering adaptation to novel control conditions.

While CtrLoRA~\cite{xu2024ctrlora} improves transferability by assigning a dedicated LoRA~\cite{hulora} to each condition during training, thereby shaping a more adaptable ControlNet, it remains computationally intensive due to large-scale multi-condition pretraining. More critically, its rigid separation mechanism fails to account for intrinsic inter-condition correlations, leading to suboptimal performance, limiting modular reuse and hindering generalization to unseen conditions.

Recently, knowledge diversion~\cite{xie2025kind} was introduced to explicitly disentangle task-agnostic and task-specific knowledge by applying Singular Value Decomposition (SVD) to factorize network weight matrices into shared learngenes and task-specific tailors, thereby improving modular reuse and cross-task transferability.
Building on this principle, we propose DivControl, which brings knowledge diversion into controllable image generation. By factorizing ControlNet into shared learngenes and lightweight tailors during training, DivControl supports unified generation across diverse conditions and enables efficient adaptation to novel conditions with minimal overhead.

DivControl applies SVD to decompose each weight matrix into basic components—pairs of singular vectors—which are modularly assigns them to shared learngenes or condition-specific tailors for structured control.
To improve parameter sharing and generalization, DivControl replaces the binary gate~\cite{xie2025kind, xu2024ctrlora} with a dynamic gate.
Specifically, each condition is described by a textual instruction, which is encoded into a condition text embedding using a pretrained text encoder~\cite{oquab2024dinov2}. 
The embedding guides the dynamic gate to assign soft weights over tailors, following a mixture-of-experts style~\cite{zhou2022mixture, riquelme2021scaling} to facilitate modular reuse and enhance zero-shot generalization to novel conditions.
To enhance convergence and semantic alignment during knowledge diversion, we incorporate a representation alignment module~\cite{yu2024representation}, which aligns condition image embeddings with shallow diffusion features.This auxiliary supervision improves early feature learning, enhances knowledge decomposition, and strengthens alignment between generated images and target conditions.

DivControl is pretrained via knowledge diversion on Subject200K~\cite{tan2024ominicontrol} using 8 base conditions and evaluated on COCO2017~\cite{lin2014microsoft} with 10 additional unseen conditions to assess generalization and transferability.
Remarkably, DivControl requires only 450K training steps ($\sim$165 GPU hours), achieving a 36.4× reduction in computational cost compared to CtrLoRA (6000 GPU hours~\cite{xu2024ctrlora}), while improving unified controllability with average CLIP-I gains of 0.05, respectively, on base conditions.
For unseen conditions, DivControl demonstrates strong zero-shot generalization, generating high-quality outputs on low-shift conditions (e.g., \textsc{Grayscale} and \textsc{Lineart}) without finetuning.
For high-shift conditions, DivControl achieves state-of-the-art performance by finetuning only the tailors at minimal cost ($\sim$0.23 GPU hours and 200 images), offering a sharp contrast to ControlNet, which requires over 600 GPU hours per condition for retraining.

Our main contributions are as follows:
1) We introduce DivControl, the first decomposable framework for controllable image generation. By factorizing ControlNet through knowledge diversion, DivControl enables modular, interpretable, and transferable generation with substantially reduced computational overhead.
2) We propose a representation alignment mechanism that bridges condition inputs and diffusion features, enhancing controllability and accelerating convergence in controllable image generation.
3) We construct a benchmark with 18 diverse control conditions to evaluate unified controllable generation, as well as transferability and generalization of trained models. Extensive experiments demonstrate that DivControl consistently outperforms prior methods across both seen and unseen conditions.

\section{Related Work}
\subsection{Controllable Image Generation}
Diffusion models have achieved significant progress in text-to-image (T2I) synthesis, producing high-fidelity images across diverse prompts~\cite{balaji2022ediff, nichol2022glide, peebles2023scalable}.
Recent efforts extend control to spatial conditions, such as depth and segmentation maps~\cite{chen2024pixart, mou2024t2i}.
ControlNet~\cite{zhang2023adding} exemplifies this by introducing separate branches, but training separate models for each condition is computationally expensive. Adapter-based alternatives~\cite{mou2024t2i, zavadski2024controlnet} reduce cost but remain data-intensive.

To enhance flexibility, recent methods explore unified controllable generation, aiming to use a single model across diverse conditions~\cite{qin2023unicontrol, zhao2023uni, hu2023cocktail}. 
However, these models struggle to generalize to novel or semantically divergent conditions.
We address this by introducing a unified framework that integrates modular parameter decomposition with dynamic routing. By disentangling condition-agnostic and condition-specific knowledge during training, our method enables scalable generation across diverse conditions and efficient adaptation to unseen conditions, even under limited resources.

\subsection{Learngene and Knowledge Diversion}
The \textsc{Learngene} framework, inspired by biological inheritance~\cite{feng2023genes, wang2023learngene}, encodes task-agnostic knowledge into modular neural units for efficient transfer~\cite{feng2024transferring}. Existing approaches mainly emphasize knowledge compression and reuse—either through heuristic layer selection~\cite{wang2022learngene, wang2023learngene} or structured operations such as Kronecker products~\cite{feng2024wave}. 
However, they largely focus on representation reuse without explicit task-level disentanglement, limiting adaptability in multi-task and cross-domain scenarios.

To address this, knowledge diversion~\cite{xie2025kind} decomposes model parameters into task-agnostic learngenes and task-specific tailors, facilitating modular reuse through gated routing. Recent works, such as FAD~\cite{shi2025fad}, apply this to the frequency domain for few-shot adaptation. 
We extend this framework to controllable image generation by disentangling ControlNet parameters across conditions, enabling unified multi-condition generation and efficient adaptation to novel conditions.

\begin{figure*}[t]
  \centering
  \includegraphics[width=\linewidth]{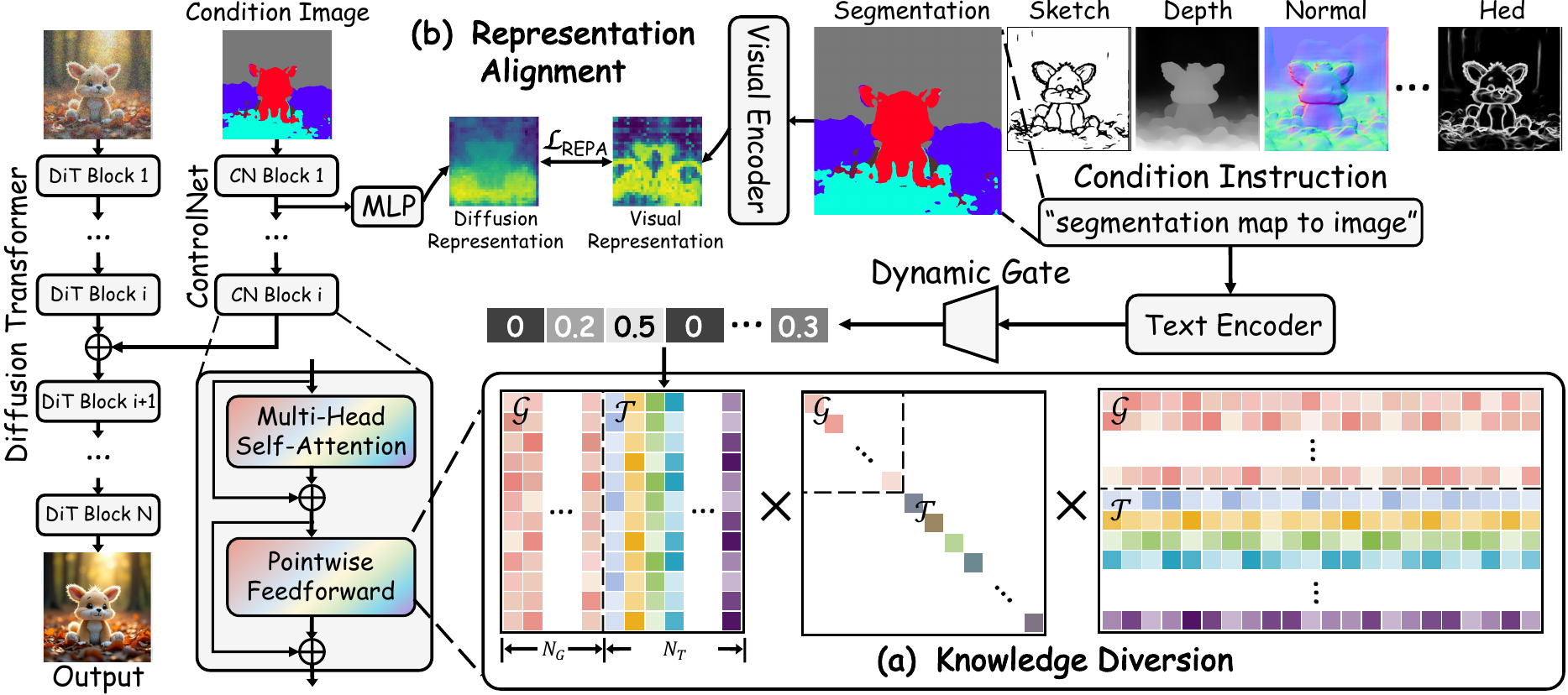}
  \vspace{-0.25in}
  \caption{\textbf{Overview of the DivControl.} (a)~Each weight matrix in ControlNet is factorized via SVD into condition-agnostic learngenes and condition-specific tailors. A dynamic gate routes each input to relevant tailors while jointly updating shared learngenes, enabling modular and disentangled representation across conditions.
  (b)~Shallow features in ControlNet are aligned with condition semantics extracted by a pre-trained vision encoder, improving consistency and accelerating convergence.}
  \label{fig:method1}
  \vspace{-0.1in}
\end{figure*}

\begin{figure}[tb]
  \centering
  \includegraphics[width=\linewidth]{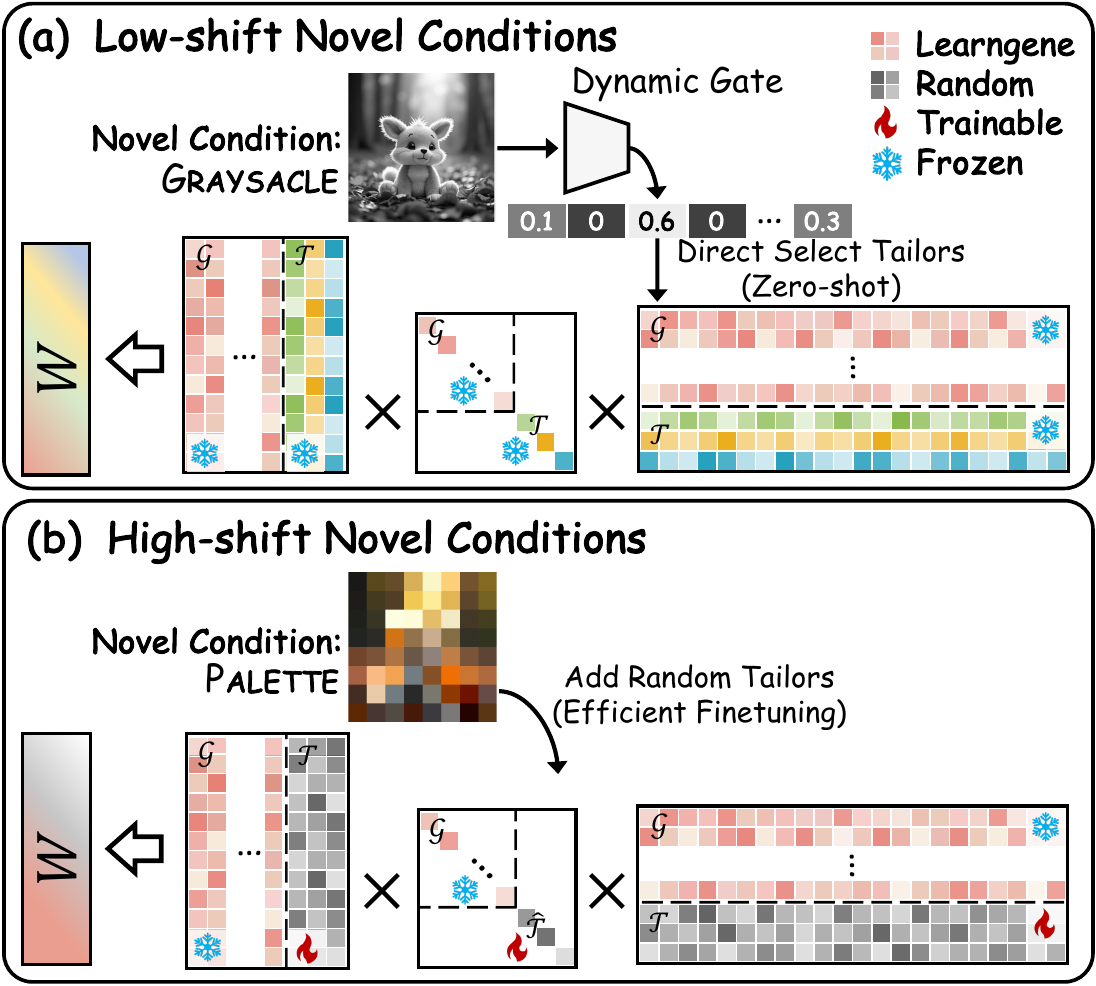}
  \vspace{-0.25in}
  \caption{\textbf{Generalization to Novel Control Conditions.} (a)~For low-shift conditions, DivControl leverages instruction embeddings to dynamically activate semantically aligned tailors, enabling zero-shot generation. (b)~For high-shift conditions, it reuses condition-agnostic learngenes while introducing randomly initialized tailors, supporting efficient few-shot adaptation.}
  \label{fig:method2}
  \vspace{-0.1in}
\end{figure}

\section{Methods}
\subsection{Preliminary}
\subsubsection{Latent Diffusion Models}
Latent diffusion models (LDMs) shift the generative process from high-dimensional pixel space to a lower-dimensional latent space. To enable this, an autoencoder $\mathcal{E}$ encodes an image $x$ into a latent representation $z = \mathcal{E}(x)$, and a diffusion model is trained to reconstruct $z$ through a denoising process, minimizing:
\begin{equation}
    \mathcal{L}_{\text{diff}} = \mathbb{E}_{z, c, \varepsilon, t} \left[ \left|| \varepsilon - \varepsilon_\theta(z_t \mid c, t) |\right|_2^2 \right],
\label{eq:loss}
\end{equation}
where $\varepsilon_\theta$ is a noise prediction network that predicts the noise $\varepsilon$ added to $z_t$ at timestep $t$, conditioned on $c$.

\subsubsection{Conditional Generation}
In conditional LDMs, the condition $c$ is derived from external modalities such as text, class labels, or condition images.
We focus on image-to-image (I2I) generation, where a condition image $x_{\text{cond}}$ (e.g., a depth map) provides structural guidance. 
A condition encoder $\mathcal{F}$ (e.g., ControlNet~\cite{zhang2023adding}) maps $x_{\text{cond}}$ to an embedding $c=\mathcal{F}(x_{\text{cond}})$, which is injected into the denoising network via adding or similar fusion strategies.

Unlike prior methods that adopt fixed~\cite{qin2023unicontrol, xu2024ctrlora} or task-specific~\cite{zavadski2024controlnet, mou2024t2i} encoders, we apply knowledge diversion to modularize 
$\mathcal{F}$. Specifically, we decompose it into condition-agnostic components (i.e., learngenes) and condition-specific components (i.e., tailors), enabling scalable and efficient adaptation to diverse conditions.

\subsection{Knowledge Diversion in ControlNet}
\subsubsection{Decomposition of ControlNet}
We adopt a Diffusion Transformer (DiT)~\cite{chen2024pixart, peebles2023scalable, esser2024scaling} as the backbone for generation, with the corresponding ControlNet $\mathcal{F}$ sharing the same transformer-based architecture, as shown in Figure~\ref{fig:method1}. To enable flexible adaptation to diverse conditions, we apply knowledge diversion within $\mathcal{F}$, a transformer-based module composed of $L$ layers, each containing repeated projection matrices in attention and MLP blocks.
The parameter set is denoted as $\theta = \{W_{q}^{(1\thicksim L)},$ $W_{k}^{(1\thicksim L)},$ $W_{v}^{(1\thicksim L)},$ $W_{o}^{(1\thicksim L)},$ $W_{in}^{(1\thicksim L)},$ $W_{out}^{(1\thicksim L)}\}$\footnote{$W_{q}^{(1\thicksim L)}$ denotes the set $\{W_{q}^{(1)}, W_{q}^{(2)}, \dots, W_{q}^{(L)}\}$. Similar notations throughout the paper follow this convention.}. 

Following the decomposition strategy of KIND~\cite{xie2025kind}, each weight matrix $W_{\star}^{(l)}$ ($\star \in \mathcal{S} = \{q, k, v, o, in, out\}$ and $l \in [1, L]$) is factorized via SVD as:
\begin{equation}
W_{\star}^{(l)} = U_{\star}^{(l)} \Sigma_{\star}^{(l)} {V_{\star}^{(l)}}^\top = \sum_{i=1}^{r} u_{\star}^{(l, i)} \sigma_{\star}^{(l, i)} v_{\star}^{(l, i)}
\label{equ:svd}
\end{equation}
where $\Theta_{\star}^{(l, i)} = (u_{\star}^{(l, i)}, \sigma_{\star}^{(l, i)}, v_{\star}^{(l, i)})$ denotes the $i$-th \mbox{rank-1} component, and $r$ is the rank of $W_{\star}^{(l)}$. Each component $\Theta_{\star}^{(l, i)}$ captures a modular unit of structured knowledge.

\subsubsection{Dynamic Diversion via Condition Text Embedding}
To support flexible adaptation across heterogeneous conditions, these components are explicitly partitioned into $N_G$ condition-agnostic \textit{\textbf{learngenes}} $\mathcal{G}$ and $N_T$ condition-specific \textit{\textbf{tailors}} $\mathcal{T}$, where $r = N_G + N_T$. Formally:
\begin{align*}
    \mathcal{G} &= \left\{ \Theta_{\star}^{(l, i)} \mid i \in [0, N_G),\ \star \in \mathcal{S},\ l \in [1, L] \right\} \\
    \mathcal{T} &= \left\{ \Theta_{\star}^{(l, i)} \mid i \in [N_G, N_G + N_T),\ \star \in \mathcal{S},\ l \in [1, L] \right\}
\end{align*}

Unlike KIND~\cite{xie2025kind}, which relies on discrete class labels for component assignment, or CtrLoRA~\cite{xu2024ctrlora}, which employs fixed task-specific adapters, we introduce a continuous diversion mechanism to enable generalization to heterogeneous and unseen conditions.
To compensate for the limited semantics in condition images, each condition is paired with a manually defined condition instruction $t_{\text{cond}}$, which is encoded into the semantic embedding $e_{\text{txt}} = E_{\text{txt}}(t_{\text{cond}})$ using a pretrained text encoder. 

To enable dynamic adaptation, the condition embedding $e_{\text{txt}}$ is processed by a lightweight gating module $G$, analogous to the router in Mixture-of-Experts architectures~\cite{zhou2022mixture, riquelme2021scaling}. This dynamic gate produces soft weights over $N_T$ tailor components:
\begin{equation}
    \alpha = \text{softmax}(G(e_{\text{txt}})) \in \mathbb{R}^{N_T}
\label{eq:router}
\end{equation}
where $\alpha$ denotes globally shared mixing coefficients, applied uniformly across all layers $l \in [1, L]$ and projection types $\star \in \mathcal{S}$. These coefficients modulate tailor components through weighted aggregation, enabling flexible and condition-aware adaptation.

During training, the condition embedding $e_{\text{cond}}$ activates relevant tailor components via a dynamic gate, while shared learngenes are updated across all conditions. 
The condition-adaptive weight matrix is constructed as a gated combination of learngenes and tailors:
\begin{equation}
    \widetilde{W}_{\star}^{(l)} = \mathcal{G}_{\star}^{(l)} + \sum_{k=1}^{K} \alpha \cdot \mathcal{T}_{k, \star}^{(l)}.
\label{eq:tailor_mix}
\end{equation}
This facilitates explicit disentanglement of condition-agnostic and condition-specific knowledge, with all components—learngenes, tailors, and the dynamic gate—jointly optimized via end-to-end training (see Algorithm~\ref{alg:algorithm} in Appendix).

\subsection{Representation Alignment}
Representation Alignment (REPA)~\cite{yu2024representation} was originally proposed to improve training efficiency and synthesis quality in class-conditioned diffusion models by aligning intermediate features with those from pretrained vision encoders~\cite{wu2025representation, tian2025u, jiang2025no, leng2025repa}. 
We extend REPA to controllable image generation, where it facilitates faster optimization of the condition encoder and improves control fidelity.

Given a condition image $x_{\text{cond}}$, we extract a semantic embedding $e_{\text{img}} = E_{\text{img}}(x_{\text{cond}}) \in \mathbb{R}^{N\times d}$ using a frozen vision encoder.
Simultaneously, a shallow feature $f_{\text{cond}} \in \mathbb{R}^{N\times d'}$ is obtained from early layers of the ControlNet $\mathcal{F}$. 
A lightweight MLP head $\mathcal{A}(\cdot)$ is trained to align $f_{\text{cond}}$ with $e_{\text{img}}$ via
\begin{equation}
    \mathcal{L}_{\text{REPA}} = - \mathbb{E}_{z, c, \varepsilon, t}\left[\frac{1}{N}\sum_{n=1}^N \text{sim}(\mathcal{A}(f_{\text{cond}})^{[n]}, e_{\text{img}}^{[n]}) \right]
\label{eq:align}
\end{equation}
where $n$ is the patch index. This alignment encourages $\mathcal{F}$ to learn semantically grounded features, improving convergence while enhancing the reliability of condition-guided generation.
The final objective combines denoising and alignment losses to optimize both generative fidelity and control semantics:
\begin{equation}
    \mathcal{L}_{\text{total}} = \mathcal{L}_{\text{diff}} + \lambda \cdot \mathcal{L}_{\text{REPA}},
\label{eq:loss2}
\end{equation}
where $\mathcal{L}_{\text{diff}}$ is the standard denoising loss (Eq.~\eqref{eq:loss}), and $\lambda$ balances the contribution of the alignment regularization.


\subsection{Efficient Adaptation to Novel Conditions}
Through knowledge diversion, DivControl factorizes ControlNet into condition-agnostic learngenes and condition-specific tailors, enabling rapid adaptation to novel conditions via reusable general features and modular activation of specialized components.

Given a low-shift novel condition $x_{\text{low}}$, we directly encode corresponding condition instruction via a pretrained text encoder to obtain $e_{\text{low}} = E_{\text{txt}}(t_{\text{low}})$, which is fed into the dynamic gate $G$ to compute soft routing weights over $N_T$ tailor components, as shown in Figure~\ref{fig:method2}a.
\begin{equation}
    \alpha_{\text{low}} = \text{softmax}(G(e_{\text{low}})).
\end{equation}
Leveraging the semantic generalization of $E_{\text{txt}}$, this mechanism enables direct zero-shot generation for unseen conditions without gradient updates or task-specific retraining.

For conditions $x_{\text{high}}$ with substantial semantic shifts (Figure~\ref{fig:method2}b), adaptation remains lightweight by introducing randomly initialized tailors while keeping transferred parameters frozen, enabling localized fine-tuning without disrupting previously acquired general knowledge.

This modular adaptation strategy promotes generalization, facilitates cross-condition transfer, and ensures efficient adaptation with minimal overhead.

\renewcommand{\arraystretch}{0.97}
\begin{table*}
    \centering
    \setlength{\tabcolsep}{1 mm}
    \caption{\textbf{Performance on Basic Control Conditions.} We report LPIPS ($\downarrow$), SSIM ($\uparrow$), and CLIP-I ($\uparrow$) across eight \textbf{\textit{basic control conditions}} to evaluate generation fidelity and semantic alignment. ``GPU Hour'' indicates total training time, and ``Para.'' denotes the average number of trainable parameters, reflecting overall training efficiency.}
    \vspace{-0.1in}
    \resizebox{0.99\textwidth}{!}{
        \begin{tabular}{@{}lllllllllllll|lll@{}}
        \toprule[1.5pt]
        & \multicolumn{3}{c}{\textsc{BBox}} & \multicolumn{3}{c}{\textsc{Canny}} & \multicolumn{3}{c}{\textsc{Depth}} & \multicolumn{3}{c|}{\textsc{Hed}} & \multicolumn{3}{c}{\textbf{\textit{Cost}}}\\ 
        \cmidrule(lr){2-4} 
        \cmidrule(lr){5-7}
        \cmidrule(lr){8-10}
        \cmidrule(lr){11-13} 
        \cmidrule(l){14-16} 
        & LPIPS & SSIM & CLIP-I & LPIPS & SSIM & CLIP-I & LPIPS & SSIM & CLIP-I & LPIPS & SSIM & CLIP-I & \multicolumn{2}{c}{GPU Hour} & Para. \\ 
        \cmidrule[1.1pt](r){1-13}
        \cmidrule[1.1pt](l){14-16}
        Pixart-$\delta$ 
        & 0.255 & 0.766 & 89.43 
        & 0.283 & 0.473 & 93.69
        & 0.232 & 0.823 & 93.20 
        & 0.246$^*$ & 0.675$^*$ & 95.00$^*$ 
        & \multicolumn{2}{c}{8$\times$36 h} & \multicolumn{1}{r}{\small{8$\times$295}M} \\
        UniControl 
        & 0.229$^*$ & 0.777 & 90.11$^*$ 
        & \textbf{0.249} & \textbf{0.500} & \textbf{94.97}
        & 0.229 & 0.837$^*$ & 93.68 
        & \textbf{0.228} & \textbf{0.704} & \textbf{95.44} 
        & \multicolumn{2}{c}{5000 h} & \textbf{374M} \\
        CtrLoRA-SD 
        & \textbf{0.221} & \textbf{0.788} & \textbf{90.50} 
        & 0.316 & 0.404 & 93.66
        & \textbf{0.218} & \textbf{0.841} & \textbf{94.25} 
        & 0.285 & 0.609 & 94.27
        & \multicolumn{2}{c}{6000 h} & 656M \\
        CtrLoRA-PA 
        & 0.249 & 0.768 & 89.62
        & 0.284 & 0.473$^*$ & 93.59
        & 0.236 & 0.812 & 93.36 
        & 0.270 & 0.638 & 94.30 
        & \multicolumn{2}{c}{\textbf{165 h}} & 519M \\
        \cmidrule(r){1-13} 
        \cmidrule(l){14-16} 
        \cellcolor{gray!15}{DivControl}
        & \cellcolor{gray!15}{0.237} & \cellcolor{gray!15}{0.779$^*$}& \cellcolor{gray!15}{89.99} 
        & \cellcolor{gray!15}{0.274$^*$} & \cellcolor{gray!15}{0.466} & \cellcolor{gray!15}{94.03$^*$}
        & \cellcolor{gray!15}{0.223$^*$} & \cellcolor{gray!15}{0.833} & \cellcolor{gray!15}{93.79$^*$} 
        & \cellcolor{gray!15}{0.259} & \cellcolor{gray!15}{0.657} & \cellcolor{gray!15}{94.55} 
        & \multicolumn{2}{c}{\cellcolor{gray!15}{\textbf{165 h}}} & \cellcolor{gray!15}{477M$^*$} \\
        \midrule[1.5pt]

        & \multicolumn{3}{c}{\textsc{Sketch}} & \multicolumn{3}{c}{\textsc{Normal}} & \multicolumn{3}{c}{\textsc{Outpainting}} & \multicolumn{3}{c|}{\textsc{Segmentation}} & \multicolumn{3}{c}{\textbf{\textit{Average}}} \\ 
        \cmidrule(lr){2-4} 
        \cmidrule(lr){5-7}
        \cmidrule(lr){8-10}
        \cmidrule(lr){11-13} 
        \cmidrule(l){14-16} 
        & LPIPS & SSIM & CLIP-I & LPIPS & SSIM & CLIP-I & LPIPS & SSIM & CLIP-I & LPIPS & SSIM & CLIP-I & LPIPS & SSIM & CLIP-I \\ 
        \cmidrule[1.1pt](r){1-13}
        \cmidrule[1.1pt](l){14-16}
        Pixart-$\delta$ 
        & 0.260 & 0.718 & 91.70 
        & 0.400 & 0.705 & 89.96
        & 0.061 & 0.898 & 93.42 
        & 0.458 & 0.652 & 87.92 
        & 0.274 & 0.714 & 91.79 \\
        UniControl
        & 0.344 & 0.628 & 87.22
        & 0.382$^*$ & \textbf{0.728} & 90.58$^*$
        & 0.066 & 0.909 & 93.17 
        & 0.461 & 0.647 & 87.61$^*$ 
        & 0.273 & 0.716$^*$ & 91.72 \\
        CtrLoRA-SD 
        & 0.256$^*$ & 0.721 & 92.08$^*$ 
        & \textbf{0.377} & 0.710$^*$ & \textbf{91.15}
        & 0.072 & 0.893 & 92.95 
        & \textbf{0.437} & \textbf{0.662} & \textbf{89.03} 
        & 0.273$^*$ & 0.703 & 92.24$^*$  \\
        CtrLoRA-PA 
        & 0.257 & 0.721$^*$ & 91.67 
        & 0.400 & 0.697 & 90.16
        & 0.058$^*$ & 0.914$^*$ & 93.80$^*$ 
        & 0.447 & 0.656$^*$ & 88.35 
        & 0.275 & 0.710 & 91.85 \\
        \cmidrule(r){1-13} 
        \cmidrule(l){14-16} 
        \cellcolor{gray!15}{DivControl}
        & \cellcolor{gray!15}{\textbf{0.242}} & \cellcolor{gray!15}{\textbf{0.741}} & \cellcolor{gray!15}{\textbf{92.25}} 
        & \cellcolor{gray!15}{0.386} & \cellcolor{gray!15}{0.710} & \cellcolor{gray!15}{90.53}
        & \cellcolor{gray!15}{\textbf{0.053}} & \cellcolor{gray!15}{\textbf{0.920}} & \cellcolor{gray!15}{\textbf{94.51}} 
        & \cellcolor{gray!15}{0.445$^*$} & \cellcolor{gray!15}{0.656} & \cellcolor{gray!15}{88.66} 
        & \cellcolor{gray!15}{\textbf{0.265}} & \cellcolor{gray!15}{\textbf{0.720}} & \cellcolor{gray!15}{\textbf{92.29}} \\
        \bottomrule[1.5pt]
        \end{tabular}
        }
    \label{tab:main}
    \vspace{-0.05in}
\end{table*}

\section{Experimental Setup}
\subsection{Dataset}
\label{sec:dataset}
We perform knowledge diversion on Subject200K~\cite{tan2024ominicontrol}, a large-scale dataset containing 200K FLUX-generated~\cite{flux} images prompted by GPT-4o~\cite{openai2023gpt4}. We annotate each image with 8 basic conditions for unified conditional generation. For evaluation, we follow CtrLoRA~\cite{xu2024ctrlora} and use the COCO2017~\cite{lin2014microsoft} validation set, extended with 10 additional conditions spanning both low- and high-shift distributions to assess DivControl’s zero-shot generalization and efficient finetuning capabilities. Annotation details and examples are provided in the Appendix~\ref{app:dataset}.

\subsection{Basic Setting}
We build on PixArt-$\delta$~\cite{chen2024pixart}, adopting a DiT backbone with a $64 \times 64$ latent resolution. The ControlNet $\mathcal{F}$ shares the same 13-layer transformer architecture. Models are trained for 450K steps using AdamW with a learning rate of $1.25 \times 10^{-5}$, weight decay of $3 \times 10^{-2}$, and batch size 16. A MultiStepLR scheduler decays the learning rate by 0.4 at 300K steps.

For knowledge diversion, we set the number of learngenes $N_G$ and tailors $N_T$ to 576, with 288 active tailor components per condition. For REPA, features from the 4-th layer of $\mathcal{F}$ are aligned with DINOv2-B~\cite{oquab2024dinov2} embeddings via an alignment loss weighted by $\lambda = 0.05$. Additional hyperparameter details are provided in Appendix~\ref{app:hyper}.

\section{Results}
\subsection{Universal Generation on Basic Conditions}
DivControl enables unified controllable generation by decomposing ControlNet into reusable learngenes and condition-specific tailors through knowledge diversion, supporting scalable multi-condition control.
As shown in Table~\ref{tab:main}, DivControl outperforms CtrLoRA~\cite{xu2024ctrlora} on all metrics in the average of eight basic conditions (Section~\ref{sec:dataset}), while reducing training time to just 165 GPU hours—much lower than the 6000 and 5000 hours required by CtrLoRA and UniControl, respectively—demonstrating both efficiency and strong generalization.

In contrast, existing methods face structural bottlenecks. UniControl~\cite{qin2023unicontrol} employs a shared encoder that lacks task-specific specialization. CtrLoRA introduces modularity through LoRA but relies on static binary gate without semantic-aware routing or representation disentanglement, limiting transferability and reuse.

DivControl mitigates these limitations through a dynamic gate, which enables semantic-aware selection of condition-specific tailors based on condition embeddings. 
By explicitly disentangling condition-agnostic learngenes from condition-specific modulations, it promotes modular reuse, accelerates convergence, and enhances scalability—highlighting the value of structured decomposition for unified controllable generation.

\begin{figure*}[h]
  \centering
  \includegraphics[width=\linewidth]{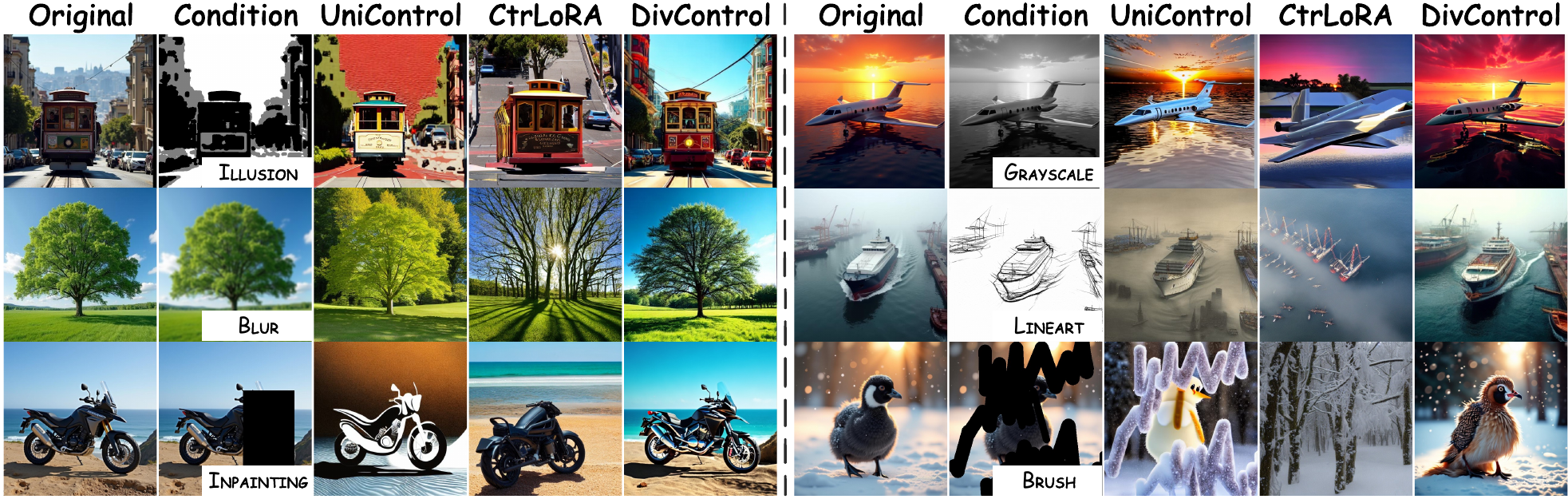}
  \vspace{-0.25in}
  \caption{\textbf{Zero-shot Generalization on Low-shift Novel Conditions.} We visualize qualitative results on unseen conditions that are semantically aligned with training conditions. DivControl achieves effective zero-shot generation by leveraging knowledge diversion and dynamic gating, which route condition inputs to semantically relevant tailor components without any fine-tuning.}
  \label{fig:zero}
  \vspace{-0.1in}
\end{figure*}

\begin{table*}
    \centering
    \setlength{\tabcolsep}{1 mm}
    \resizebox{0.99\textwidth}{!}{
        \begin{tabular}{@{}lllllllllllll|lll@{}}
        \toprule[1.5pt]
        & \multicolumn{3}{c}{\textsc{Blur}} & \multicolumn{3}{c}{\textsc{Brush}} & \multicolumn{3}{c}{\textsc{Grayscale}} & \multicolumn{3}{c|}{\textsc{Inpainting}} & \multicolumn{3}{c}{\textbf{\textit{Cost}}}\\ 
        \cmidrule(lr){2-4} 
        \cmidrule(lr){5-7}
        \cmidrule(lr){8-10}
        \cmidrule(lr){11-13} 
        \cmidrule(l){14-16} 
        & LPIPS & SSIM & CLIP-I & LPIPS & SSIM & CLIP-I & LPIPS & SSIM & CLIP-I & LPIPS & SSIM & CLIP-I & \multicolumn{2}{c}{GPU Time} & Para. \\ 
        \cmidrule[1.1pt](r){1-13}
        \cmidrule[1.1pt](l){14-16}
        Pixart-$\delta$ 
        & 0.179 & 0.923 & 94.70
        & 0.168 & 0.769 & 94.56
        & 0.209 & 0.708 & 96.92
        & 0.149$^*$ & 0.771 & 96.45
        & \multicolumn{2}{c}{1.01} & 295M \\
        Pixart-$\delta$-canny 
        & 0.162$^*$ & 0.933$^*$ & 95.20$^*$
        & 0.148$^*$ & 0.792$^*$ & 95.65$^*$
        & 0.195$^*$ & 0.723$^*$ & 97.18$^*$
        & 0.160 & 0.771$^*$ & 96.46
        & \multicolumn{2}{c}{1.01} & 295M \\
        CtrLoRA-SD 
        & 0.204 & 0.875 & 93.79
        & 0.167 & 0.729 & 95.15
        & 0.262 & 0.599 & 95.49
        & 0.197 & 0.697 & 95.34
        & \multicolumn{2}{c}{0.93} & 37M \\
        CtrLoRA-PA 
        & 0.171 & 0.927 & 94.95
        & 0.158 & 0.777 & 95.60
        & 0.201 & 0.703 & 97.13
        & 0.160 & 0.767 & 96.49$^*$
        & \multicolumn{2}{c}{\textbf{0.23}} & \textbf{14M} \\
        \cmidrule(r){1-13} 
        \cmidrule(l){14-16} 
        \cellcolor{gray!15}{DivControl}
        & \cellcolor{gray!15}{\textbf{0.143}} & \cellcolor{gray!15}{\textbf{0.946}} & \cellcolor{gray!15}{\textbf{95.84}}
        & \cellcolor{gray!15}{\textbf{0.136}} & \cellcolor{gray!15}{\textbf{0.800}} & \cellcolor{gray!15}{\textbf{96.28}}
        & \cellcolor{gray!15}{\textbf{0.195}} & \cellcolor{gray!15}{\textbf{0.724}} & \cellcolor{gray!15}{\textbf{97.33}}
        & \cellcolor{gray!15}{\textbf{0.144}} & \cellcolor{gray!15}{\textbf{0.778}} & \cellcolor{gray!15}{\textbf{96.97}}
        & \multicolumn{2}{c}{\cellcolor{gray!15}{\textbf{0.23}}} & \cellcolor{gray!15}{\textbf{14M}} \\
        \midrule[1.5pt]

        & \multicolumn{3}{c}{\textsc{Jpeg}} & \multicolumn{3}{c}{\textsc{Palette}} & \multicolumn{3}{c}{\textsc{Pixel}} & \multicolumn{3}{c|}{\textsc{Shuffle}} & \multicolumn{3}{c}{\textbf{\textit{Average}}} \\ 
        \cmidrule(lr){2-4} 
        \cmidrule(lr){5-7}
        \cmidrule(lr){8-10}
        \cmidrule(lr){11-13} 
        \cmidrule(l){14-16} 
        & LPIPS & SSIM & CLIP-I & LPIPS & SSIM & CLIP-I & LPIPS & SSIM & CLIP-I & LPIPS & SSIM & CLIP-I & LPIPS & SSIM & CLIP-I \\ 
        \cmidrule[1.1pt](r){1-13}
        \cmidrule[1.1pt](l){14-16}
        Pixart-$\delta$ 
        & 0.308 & 0.630 & 92.18
        & 0.254 & 0.675 & 88.54
        & 0.390 & 0.576 & 87.42
        & 0.685 & 0.215 & \textbf{84.80}
        & 0.293 & 0.658 & 91.95 \\
        Pixart-$\delta$-canny
        & 0.300$^*$ & 0.656$^*$ & 92.92$^*$
        & 0.234 & 0.724$^*$ & 88.89
        & 0.366$^*$ & 0.616$^*$ & 89.34$^*$
        & 0.679$^*$ & 0.229 & 84.48
        & 0.281$^*$ & 0.681$^*$ & 92.51$^*$ \\
        CtrLoRA-SD 
        & 0.356 & 0.570 & 92.07
        & 0.285 & 0.663 & 87.48
        & 0.465 & 0.500 & 86.37
        & 0.692 & 0.196 & 84.74$^*$
        & 0.328 & 0.603 & 91.30 \\
        CtrLoRA-PA 
        & 0.328 & 0.635 & 92.67 
        & 0.228$^*$ & 0.721 & 89.24$^*$
        & 0.383 & 0.588 & 88.76 
        & 0.681 & 0.237$^*$ & 83.84 
        & 0.289 & 0.669 & 92.34 \\
        \cmidrule(r){1-13} 
        \cmidrule(l){14-16} 
        \cellcolor{gray!15}{DivControl}
        & \cellcolor{gray!15}{\textbf{0.297}} & \cellcolor{gray!15}{\textbf{0.668}} & \cellcolor{gray!15}{\textbf{94.23}}
        & \cellcolor{gray!15}{\textbf{0.224}} & \cellcolor{gray!15}{\textbf{0.736}} & \cellcolor{gray!15}{\textbf{89.30}}
        & \cellcolor{gray!15}{\textbf{0.365}} & \cellcolor{gray!15}{\textbf{0.640}} & \cellcolor{gray!15}{\textbf{89.84}}
        & \cellcolor{gray!15}{\textbf{0.674}} & \cellcolor{gray!15}{\textbf{0.244}} & \cellcolor{gray!15}{84.74}
        & \cellcolor{gray!15}{\textbf{0.272}} & \cellcolor{gray!15}{\textbf{0.692}} & \cellcolor{gray!15}{\textbf{93.07}} \\
        \bottomrule[1.5pt]
        \end{tabular}
        }
    \vspace{-0.1in}
    \caption{\textbf{Performance on Novel Control Conditions.} We report LPIPS ($\downarrow$), SSIM ($\uparrow$), and CLIP-I ($\uparrow$) across 6 \textbf{\textit{high-shift}} novel conditions and 2 low-shift novel conditions to evaluate generation fidelity and semantic alignment. ``GPU Hour'' indicates total training time, and ``Para.'' denotes the average number of trainable parameters.}
    \label{tab:novel}
    \vspace{-0.05in}
\end{table*}

\subsection{Generalization to Novel Control Conditions}
We evaluate DivControl’s generalization to unseen conditions, categorized as: (1) \textit{low-shift} conditions that are semantically close to training modalities, and (2) \textit{high-shift} conditions with substantial domain or modality gaps.

Leveraging knowledge diversion and dynamic gate, DivControl enables zero-shot generation in low-shift settings and lightweight adaptation in high-shift scenarios. We present analyses for both cases below.
\subsubsection{Zero-shot Generalization on Low-shift Conditions}
For novel conditions with minor semantic or structural deviations from training conditions, DivControl enables zero-shot generalization by embedding condition instructions into task representations that guide the dynamic gate to softly activate semantically aligned tailors. 
As shown in Figure~\ref{fig:zero}, DivControl consistently produces high-fidelity, condition-aligned images across all low-shift novel conditions without requiring gradient updates.

In contrast, CtrLoRA~\cite{xu2024ctrlora} only assigns base ControlNet to each new condition, lacking a mechanism for semantic transfer and thus failing at zero-shot adaptation. UniControl~\cite{qin2023unicontrol} manually composes base-condition combinations for novel tasks, limiting flexibility and scalability.
DivControl overcomes these limitations through input-conditioned routing over dynamically selected tailors, enabling modular reuse and scalable generalization to semantically related conditions.

\subsubsection{Efficient Finetuning on High-shift Conditions}
For novel conditions with substantial distributional shifts, such as \textsc{Palette} and \textsc{Shuffle}, we transfer the parameters and initialize tailors from scratch, as these cases exhibit significant visual discrepancies and demand distinct generative priors.

As shown in Table~\ref{tab:novel}, despite substantial distribution shifts, DivControl achieves competitive performance after just 3K finetuning steps ($\sim$0.23 GPU hours) on 200 images, improving average CLIP-I by 1.72 points over CtrLoRA. This demonstrates its ability to rapidly adapt to novel conditions with minimal computational overhead.
Furthermore, unlike full-model retraining which ignores transferability and incurs substantial cost in both training time and parameters per condition, DivControl enables rapid, lightweight adaptation while preserving fidelity and consistency.

These results highlight the scalability and flexibility of DivControl, enabled by its structural decoupling of condition-agnostic and condition-specific knowledge. This modular design supports adaptive parameter transfer for diverse downstream demands, making it well suited for open-world scenarios with evolving or unseen conditions.

\begin{figure*}[tb]
  \centering
  \includegraphics[width=\linewidth]{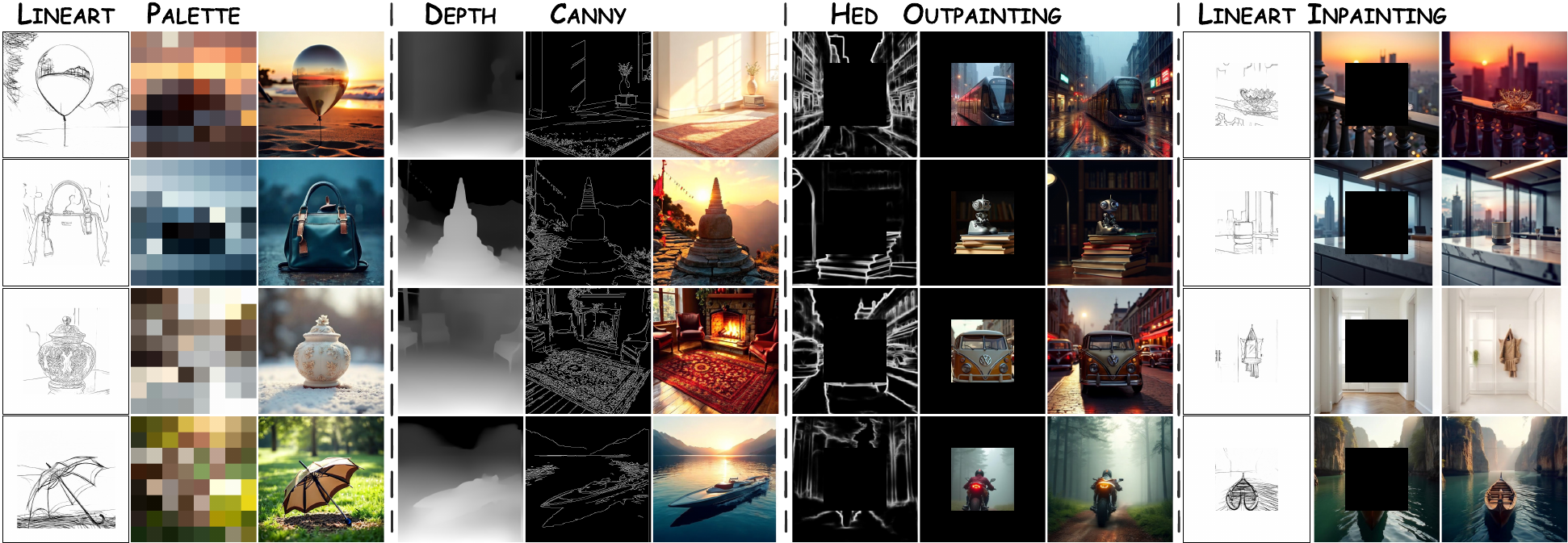}
  \vspace{-0.25in}
  \caption{\textbf{Multi-Conditional Controllable Image Generation.} 
  DivControl leverages knowledge diversion to encapsulate condition-specific knowledge into tailors, enabling flexible composition of multiple conditions. 
  This facilitates high-fidelity, semantically aligned generation under simultaneous multi-condition guidance.}
  \label{fig:multi}
  \vspace{-0.1in}
\end{figure*}

\subsection{Multi-Conditional Controllable Generation}
DivControl enables unified controllable image generation by disentangling condition-specific knowledge into independently activatable tailors, which further supports flexible composition of multiple control conditions through selective feature reuse—enabling more challenging, fine-grained controllable generation.

By aggregating task embeddings from multiple conditions, DivControl activates the corresponding tailor modules, enabling seamless integration of complex controls such as \textsc{Lineart} and \textsc{Palette}.
As shown in Figure~\ref{fig:multi}, DivControl produces high-quality and semantically consistent results even under complex, combined control conditions. This demonstrates that modular routing effectively preserves the guidance from each input while allowing them to work together for coherent generation.

\subsection{Ablation and Analysis}
\subsubsection{Ablation Experiments}
To evaluate the impact of knowledge diversion and REPA, we ablate each component individually. As shown in Table~\ref{tab:ablation}, introducing knowledge diversion alone yields notable gains, with LPIPS decreasing by 0.026, SSIM increasing by 0.03, and CLIP-I improving by 1.20\%, highlighting the effectiveness of modularizing condition-agnostic and condition-specific knowledge for improved generalization.

Incorporating REPA on top yields additional gains, with LPIPS further reduced by 0.012 and continued improvements in SSIM and CLIP-I, confirming its effectiveness in enhancing semantic alignment. 
These results underscore the complementary roles of both components: knowledge diversion enables flexible adaptation, while REPA reinforces semantic grounding—together facilitating high-quality, controllable generation across diverse conditions.
\begin{table}
    \centering
    \setlength{\tabcolsep}{1.5 mm}
    \caption{Ablation study on DivControl.}
    \vspace{-0.1in}
        \resizebox{0.48\textwidth}{!}{
        \begin{tabular}{@{}cccccc@{}}
        \toprule[1.5pt]
         & Diversion & REPA & LPIPS$\downarrow$ & SSIM$\uparrow$ & CLIP-I$\uparrow$\\
        \midrule[1.1pt]
        \#1 & & & 0.328 & 0.663 & 89.43\\
        \#2 &  \checkmark & & 0.302 & 0.693 & 90.63\\
        \midrule
        \cellcolor{gray!15}{DivControl} & \cellcolor{gray!15}{\checkmark} & \cellcolor{gray!15}{\checkmark} & \cellcolor{gray!15}{\textbf{0.290}} & \cellcolor{gray!15}{\textbf{0.696}} & \cellcolor{gray!15}{\textbf{91.31}} \\
        \bottomrule[1.5pt]
        \end{tabular}
        }
    \label{tab:ablation}
\end{table}

\subsubsection{Analysis on Dynamic Gate}
\begin{figure}[tb]
  \centering
  \includegraphics[width=\linewidth]{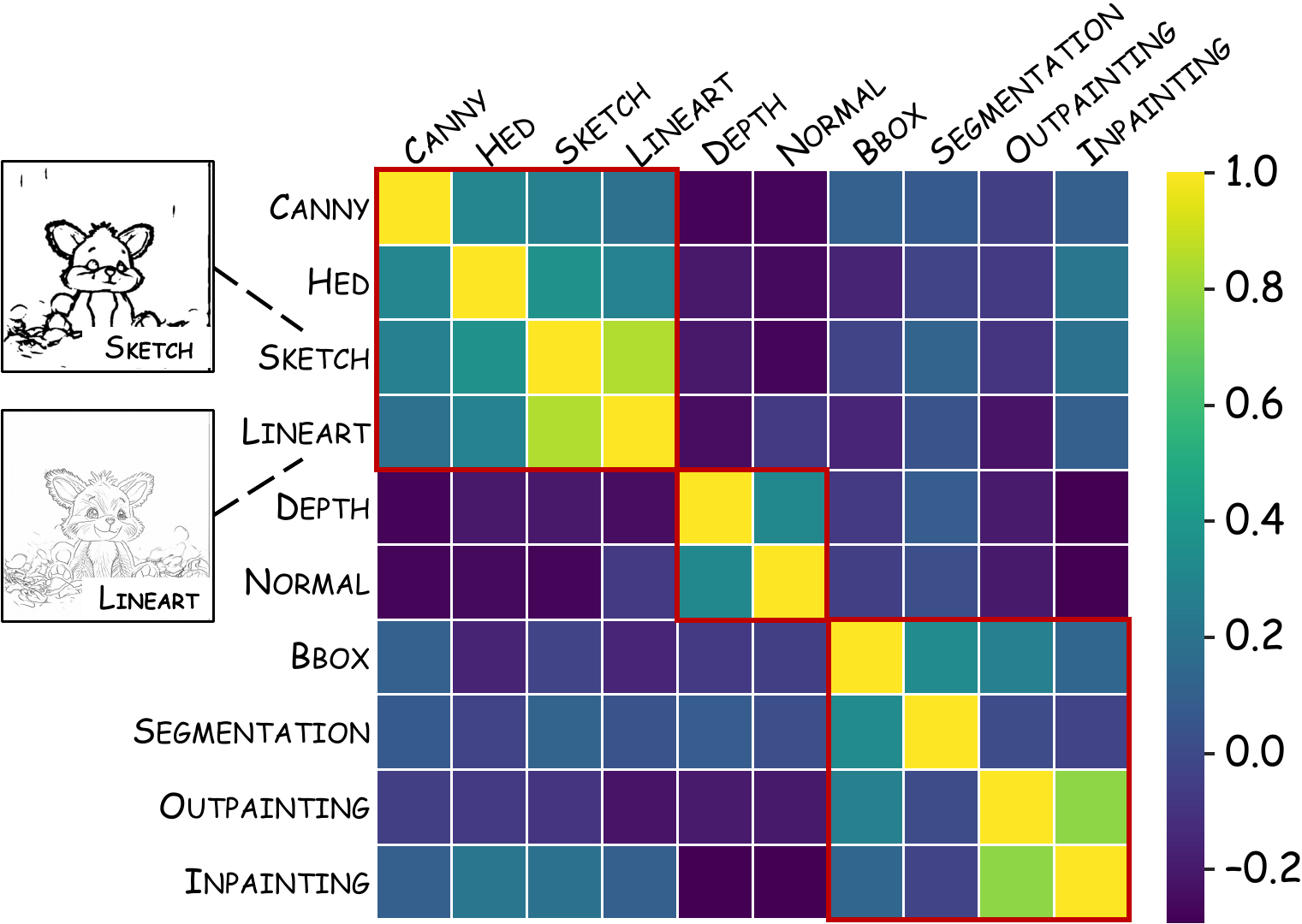}
  \vspace{-0.15in}
  \caption{Inter-condition similarity derived from dynamic gate activations ($\alpha$ in Eq.~\eqref{eq:router}), illustrating the gate’s capacity to capture semantic relationships among conditions.}
  \label{fig:gate}
  \vspace{-0.1in}
\end{figure}

To analyze the routing behavior of the dynamic gate, we visualize activation patterns across conditions. As shown in Figure~\ref{fig:gate}, semantically related conditions (e.g., \textsc{sketch} and \textsc{lineart}) yield similar tailor activations, whereas dissimilar ones (e.g., \textsc{sketch} vs. \textsc{outpainting}) exhibit divergent routing.

These results indicate that the dynamic gate effectively captures semantic similarities between conditions and modulates tailor activation accordingly. 
This behavior underpins DivControl’s strong zero-shot performance by enabling knowledge reuse for semantically related yet unseen conditions without retraining. 
Moreover, the emergence of such alignment from condition instructions alone validates their effectiveness as condition representations and underscores the semantic sensitivity of the gating mechanism.

\subsubsection{Analysis on Position and Weight of REPA}
\begin{table}
    \centering
    \setlength{\tabcolsep}{0.9 mm}
    \caption{Impact of alignment depth and weight (i.e., $\lambda$ in Eq.~\eqref{eq:loss2}) in REPA on convergence speed and stability.}
    \vspace{-0.1in}
    \resizebox{0.485\textwidth}{!}{
        \begin{tabular}{@{}cc|cccccc@{}}
        \toprule[1.5pt]
        Depth & Weight & LPIPS$\downarrow$ & SSIM$\uparrow$ & CLIP-I$\uparrow$ & MSE$\downarrow$ & FID$\downarrow$ & FDD$\downarrow$ \\ 
        \midrule
        2 & 0.2 & 0.301 & 0.685 & 90.73 & 40.22 & 10.44 & 0.046\\
        6 & 0.2 & 0.303 & 0.682 & 90.70 & 40.34 & 10.44 & 0.046\\
        \midrule
        4 & 0.005 & 0.307 & 0.681 & 90.61 & 40.47 & 10.73 & 0.046\\
        4 & 0.5 & 0.315 & 0.672 & 89.98 & 40.76 & 11.01 & 0.053\\
        \midrule
        \cellcolor{gray!15}{4} & \cellcolor{gray!15}{0.05} & \cellcolor{gray!15}{\textbf{0.290}} & \cellcolor{gray!15}{\textbf{0.696}} & \cellcolor{gray!15}{\textbf{91.31}} & \cellcolor{gray!15}{\textbf{39.84}} & \cellcolor{gray!15}{\textbf{9.73}} & \cellcolor{gray!15}{\textbf{0.040}}\\
        \bottomrule[1.5pt]
        \end{tabular}
        }
    \label{tab:ana_repa}
    \vspace{-0.05in}
\end{table}

We analyze the impact of REPA’s alignment depth and loss weight $\lambda$ on training convergence. 
As shown in Table~\ref{tab:ana_repa}, applying REPA at an intermediate depth (layer 4) yields the fastest convergence by balancing early-stage semantic guidance and late-stage task-specific adaptation. In contrast, shallow alignment lacks semantic expressiveness, while deeper alignment introduces condition-specific noise that impairs stability.

Additionally, moderate alignment strength ($\lambda = 0.05$) provides effective regularization without overly constraining feature learning. Excessive weighting hampers adaptability, while insufficient weighting weakens alignment signals. These results highlight the importance of carefully configuring REPA to ensure efficient and stable optimization.

\section{Conclusion}
In this work, we introduce DivControl, a novel training framework for ControlNet that performs knowledge diversion to construct a modular architecture during pretraining. 
By disentangling condition-agnostic knowledge into reusable learngenes and encoding condition-specific knowledge into lightweight tailors, DivControl supports dynamic, condition-aware assembly across diverse conditions.
DivControl supports zero-shot generalization to semantically related conditions via a dynamic gate, and allows efficient adaptation to large distribution shifts through plug-in randomly initialized tailors.
Extensive experiments demonstrate that DivControl consistently outperforms existing methods while substantially reducing computational overhead.

\section*{Acknowledgement}
We sincerely appreciate Freepik for contributing to the figure design. 
This research was supported by the Jiangsu Science Foundation (BK20243012, BG2024036, BK20230832), the National Science Foundation of China (62125602, U24A20324, 92464301, 62306073), China Postdoctoral Science Foundation (2022M720028), and the Xplorer Prize.

{
    \small
    \bibliographystyle{ieeenat_fullname}
    \bibliography{main}

\begin{thebibliography}{36}
\providecommand{\natexlab}[1]{#1}
\providecommand{\url}[1]{\texttt{#1}}
\expandafter\ifx\csname urlstyle\endcsname\relax
  \providecommand{\doi}[1]{doi: #1}\else
  \providecommand{\doi}{doi: \begingroup \urlstyle{rm}\Url}\fi

\bibitem[Balaji et~al.(2022)Balaji, Nah, Huang, Vahdat, Song, Zhang, Kreis,
  Aittala, Aila, Laine, et~al.]{balaji2022ediff}
Yogesh Balaji, Seungjun Nah, Xun Huang, Arash Vahdat, Jiaming Song, Qinsheng
  Zhang, Karsten Kreis, Miika Aittala, Timo Aila, Samuli Laine, et~al.
\newblock ediff-i: Text-to-image diffusion models with an ensemble of expert
  denoisers.
\newblock \emph{arXiv preprint arXiv:2211.01324}, 2022.

\bibitem[Chen et~al.()Chen, Luo, and Xie]{chen2024pixart}
Junsong Chen, Simian Luo, and Enze Xie.
\newblock Pixart-$\delta$: Fast and controllable image generation with latent
  consistency models.
\newblock In \emph{ICML 2024 Workshop on Theoretical Foundations of Foundation
  Models}.

\bibitem[Esser et~al.(2024)Esser, Kulal, Blattmann, Entezari, M{\"u}ller,
  Saini, Levi, Lorenz, Sauer, Boesel, et~al.]{esser2024scaling}
Patrick Esser, Sumith Kulal, Andreas Blattmann, Rahim Entezari, Jonas
  M{\"u}ller, Harry Saini, Yam Levi, Dominik Lorenz, Axel Sauer, Frederic
  Boesel, et~al.
\newblock Scaling rectified flow transformers for high-resolution image
  synthesis.
\newblock In \emph{Proceedings of International Conference on Machine Learning
  (ICML'24)}, pages 1--13, 2024.

\bibitem[Feng et~al.(2023)Feng, Wang, Zhang, Li, Yang, and Geng]{feng2023genes}
Fu Feng, Jing Wang, Congzhi Zhang, Wenqian Li, Xu Yang, and Xin Geng.
\newblock Genes in intelligent agents.
\newblock \emph{arXiv preprint arXiv:2306.10225}, 2023.

\bibitem[Feng et~al.(2024)Feng, Wang, and Geng]{feng2024transferring}
Fu Feng, Jing Wang, and Xin Geng.
\newblock Transferring core knowledge via learngenes.
\newblock \emph{arXiv preprint arXiv:2401.08139}, 2024.

\bibitem[Feng et~al.(2025)Feng, Xie, Wang, and Geng]{feng2024wave}
Fu Feng, Yucheng Xie, Jing Wang, and Xin Geng.
\newblock Wave: Weight template for adaptive initialization of variable-sized
  models.
\newblock In \emph{Proceedings of the IEEE/CVF Conference on Computer Vision
  and Pattern Recognition (CVPR'25)}, pages 1--10, 2025.

\bibitem[Hu et~al.(2022)Hu, Wallis, Allen-Zhu, Li, Wang, Wang, Chen,
  et~al.]{hulora}
Edward~J Hu, Phillip Wallis, Zeyuan Allen-Zhu, Yuanzhi Li, Shean Wang, Lu Wang,
  Weizhu Chen, et~al.
\newblock Lora: Low-rank adaptation of large language models.
\newblock In \emph{Proceedings of the International Conference on Learning
  Representations (ICLR'22)}, pages 1--13, 2022.

\bibitem[Hu et~al.(2023)Hu, Zheng, Liu, Zheng, Wang, Tao, and
  Cham]{hu2023cocktail}
Minghui Hu, Jianbin Zheng, Daqing Liu, Chuanxia Zheng, Chaoyue Wang, Dacheng
  Tao, and Tat-Jen Cham.
\newblock Cocktail: Mixing multi-modality control for text-conditional image
  generation.
\newblock In \emph{Thirty-seventh Conference on Neural Information Processing
  Systems}, 2023.

\bibitem[Jiang et~al.(2025)Jiang, Wang, Li, Zhang, Wang, Wei, Dai, Zhang, and
  Wang]{jiang2025no}
Dengyang Jiang, Mengmeng Wang, Liuzhuozheng Li, Lei Zhang, Haoyu Wang, Wei Wei,
  Guang Dai, Yanning Zhang, and Jingdong Wang.
\newblock No other representation component is needed: Diffusion transformers
  can provide representation guidance by themselves.
\newblock \emph{arXiv preprint arXiv:2505.02831}, 2025.

\bibitem[Leng et~al.(2025)Leng, Singh, Hou, Xing, Xie, and Zheng]{leng2025repa}
Xingjian Leng, Jaskirat Singh, Yunzhong Hou, Zhenchang Xing, Saining Xie, and
  Liang Zheng.
\newblock Repa-e: Unlocking vae for end-to-end tuning with latent diffusion
  transformers.
\newblock \emph{arXiv preprint arXiv:2504.10483}, 2025.

\bibitem[Lin et~al.(2014)Lin, Maire, Belongie, Hays, Perona, Ramanan,
  Doll{\'a}r, and Zitnick]{lin2014microsoft}
Tsung-Yi Lin, Michael Maire, Serge Belongie, James Hays, Pietro Perona, Deva
  Ramanan, Piotr Doll{\'a}r, and C~Lawrence Zitnick.
\newblock Microsoft coco: Common objects in context.
\newblock In \emph{Computer vision--ECCV 2014: 13th European conference,
  zurich, Switzerland, September 6-12, 2014, proceedings, part v 13}, pages
  740--755. Springer, 2014.

\bibitem[{Midjourney}(2022)]{midjourney}
{Midjourney}.
\newblock Midjourney.com.
\newblock \url{https://www.midjourney.com}, 2022.
\newblock Accessed: 2024-11-14.

\bibitem[Mou et~al.(2024)Mou, Wang, Xie, Wu, Zhang, Qi, and Shan]{mou2024t2i}
Chong Mou, Xintao Wang, Liangbin Xie, Yanze Wu, Jian Zhang, Zhongang Qi, and
  Ying Shan.
\newblock T2i-adapter: Learning adapters to dig out more controllable ability
  for text-to-image diffusion models.
\newblock In \emph{Proceedings of the AAAI conference on artificial
  intelligence}, pages 4296--4304, 2024.

\bibitem[Nichol et~al.(2022)Nichol, Dhariwal, Ramesh, Shyam, Mishkin, Mcgrew,
  Sutskever, and Chen]{nichol2022glide}
Alexander~Quinn Nichol, Prafulla Dhariwal, Aditya Ramesh, Pranav Shyam, Pamela
  Mishkin, Bob Mcgrew, Ilya Sutskever, and Mark Chen.
\newblock Glide: Towards photorealistic image generation and editing with
  text-guided diffusion models.
\newblock In \emph{International Conference on Machine Learning}, pages
  16784--16804. PMLR, 2022.

\bibitem[{OpenAI}(2023)]{openai2023gpt4}
{OpenAI}.
\newblock Gpt-4: Openai language model.
\newblock \url{https://openai.com/research/gpt-4}, 2023.

\bibitem[Oquab et~al.(2024)Oquab, Darcet, Moutakanni, Vo, Szafraniec, Khalidov,
  Fernandez, Haziza, Massa, El-Nouby, et~al.]{oquab2024dinov2}
Maxime Oquab, Timoth{\'e}e Darcet, Th{\'e}o Moutakanni, Huy Vo, Marc
  Szafraniec, Vasil Khalidov, Pierre Fernandez, Daniel Haziza, Francisco Massa,
  Alaaeldin El-Nouby, et~al.
\newblock Dinov2: Learning robust visual features without supervision.
\newblock \emph{Transactions on Machine Learning Research Journal}, 2024.

\bibitem[Peebles and Xie(2023)]{peebles2023scalable}
William Peebles and Saining Xie.
\newblock Scalable diffusion models with transformers.
\newblock In \emph{Proceedings of the IEEE/CVF International Conference on
  Computer Vision (ICCV'23)}, pages 4195--4205, 2023.

\bibitem[Qin et~al.(2023)Qin, Zhang, Yu, Feng, Yang, Zhou, Wang, Niebles,
  Xiong, Savarese, et~al.]{qin2023unicontrol}
Can Qin, Shu Zhang, Ning Yu, Yihao Feng, Xinyi Yang, Yingbo Zhou, Huan Wang,
  Juan~Carlos Niebles, Caiming Xiong, Silvio Savarese, et~al.
\newblock Unicontrol: A unified diffusion model for controllable visual
  generation in the wild.
\newblock In \emph{NeurIPS}, 2023.

\bibitem[Ramesh et~al.(2022)Ramesh, Dhariwal, Nichol, Chu, and
  Chen]{ramesh2022hierarchical}
Aditya Ramesh, Prafulla Dhariwal, Alex Nichol, Casey Chu, and Mark Chen.
\newblock Hierarchical text-conditional image generation with clip latents.
\newblock \emph{arXiv preprint arXiv:2204.06125}, 1\penalty0 (2):\penalty0 3,
  2022.

\bibitem[Riquelme et~al.(2021)Riquelme, Puigcerver, Mustafa, Neumann, Jenatton,
  Susano~Pinto, Keysers, and Houlsby]{riquelme2021scaling}
Carlos Riquelme, Joan Puigcerver, Basil Mustafa, Maxim Neumann, Rodolphe
  Jenatton, Andr{\'e} Susano~Pinto, Daniel Keysers, and Neil Houlsby.
\newblock Scaling vision with sparse mixture of experts.
\newblock \emph{Advances in Neural Information Processing Systems},
  34:\penalty0 8583--8595, 2021.

\bibitem[{Shakker Labs}(2024)]{flux}
{Shakker Labs}.
\newblock Flux.1-dev-controlnet-union-pro.
\newblock
  \url{https://huggingface.co/Shakker-Labs/FLUX.1-dev-ControlNet-Union-Pro},
  2024.

\bibitem[Shi et~al.(2025)Shi, Feng, Xie, Wang, and Geng]{shi2025fad}
Ruixiao Shi, Fu Feng, Yucheng Xie, Jing Wang, and Xin Geng.
\newblock Fad: Frequency adaptation and diversion for cross-domain few-shot
  learning.
\newblock \emph{arXiv preprint arXiv:2505.08349}, 2025.

\bibitem[Tan et~al.(2024)Tan, Liu, Yang, Xue, and Wang]{tan2024ominicontrol}
Zhenxiong Tan, Songhua Liu, Xingyi Yang, Qiaochu Xue, and Xinchao Wang.
\newblock Ominicontrol: Minimal and universal control for diffusion
  transformer.
\newblock \emph{arXiv preprint arXiv:2411.15098}, 2024.

\bibitem[Tian et~al.(2025)Tian, Chen, Zheng, Liang, Xu, and Wang]{tian2025u}
Yuchuan Tian, Hanting Chen, Mengyu Zheng, Yuchen Liang, Chao Xu, and Yunhe
  Wang.
\newblock U-repa: Aligning diffusion u-nets to vits.
\newblock \emph{arXiv preprint arXiv:2503.18414}, 2025.

\bibitem[Wang et~al.(2025)Wang, Peng, He, Yang, Jin, Wu, Hu, Pan, Gan, Chi,
  et~al.]{wang2025unicombine}
Haoxuan Wang, Jinlong Peng, Qingdong He, Hao Yang, Ying Jin, Jiafu Wu, Xiaobin
  Hu, Yanjie Pan, Zhenye Gan, Mingmin Chi, et~al.
\newblock Unicombine: Unified multi-conditional combination with diffusion
  transformer.
\newblock \emph{arXiv preprint arXiv:2503.09277}, 2025.

\bibitem[Wang et~al.(2024)Wang, Gao, Zhao, Sun, and Dai]{wang2024auxiliary}
Lean Wang, Huazuo Gao, Chenggang Zhao, Xu Sun, and Damai Dai.
\newblock Auxiliary-loss-free load balancing strategy for mixture-of-experts.
\newblock \emph{arXiv preprint arXiv:2408.15664}, 2024.

\bibitem[Wang et~al.(2022)Wang, Geng, Lin, Xia, Qi, and Xu]{wang2022learngene}
QiuFeng Wang, Xin Geng, ShuXia Lin, Shi-Yu Xia, Lei Qi, and Ning Xu.
\newblock Learngene: From open-world to your learning task.
\newblock In \emph{Proceedings of the AAAI Conference on Artificial
  Intelligence (AAAI'22)}, pages 8557--8565, 2022.

\bibitem[Wang et~al.(2023)Wang, Yang, Lin, and Geng]{wang2023learngene}
Qiufeng Wang, Xu Yang, Shuxia Lin, and Xin Geng.
\newblock Learngene: Inheriting condensed knowledge from the ancestry model to
  descendant models.
\newblock \emph{arXiv preprint arXiv:2305.02279}, 2023.

\bibitem[Wu et~al.(2025)Wu, Zhang, Shi, Gao, Chen, Wang, Chen, Gao, Tang, Yang,
  et~al.]{wu2025representation}
Ge Wu, Shen Zhang, Ruijing Shi, Shanghua Gao, Zhenyuan Chen, Lei Wang, Zhaowei
  Chen, Hongcheng Gao, Yao Tang, Jian Yang, et~al.
\newblock Representation entanglement for generation: Training diffusion
  transformers is much easier than you think.
\newblock \emph{arXiv preprint arXiv:2507.01467}, 2025.

\bibitem[Xie et~al.(2025)Xie, Feng, Shi, Wang, Rui, and Geng]{xie2025kind}
Yucheng Xie, Fu Feng, Ruixiao Shi, Jing Wang, Yong Rui, and Xin Geng.
\newblock Kind: Knowledge integration and diversion for training decomposable
  models.
\newblock In \emph{Forty-second International Conference on Machine Learning},
  2025.

\bibitem[Xu et~al.(2024)Xu, He, Shan, and Chen]{xu2024ctrlora}
Yifeng Xu, Zhenliang He, Shiguang Shan, and Xilin Chen.
\newblock Ctrlora: An extensible and efficient framework for controllable image
  generation.
\newblock \emph{arXiv preprint arXiv:2410.09400}, 2024.

\bibitem[Yu et~al.(2024)Yu, Kwak, Jang, Jeong, Huang, Shin, and
  Xie]{yu2024representation}
Sihyun Yu, Sangkyung Kwak, Huiwon Jang, Jongheon Jeong, Jonathan Huang, Jinwoo
  Shin, and Saining Xie.
\newblock Representation alignment for generation: Training diffusion
  transformers is easier than you think.
\newblock In \emph{Proceedings of the International Conference on Learning
  Representations (ICLR'25)}, pages 1--17, 2024.

\bibitem[Zavadski et~al.(2024)Zavadski, Feiden, and
  Rother]{zavadski2024controlnet}
Denis Zavadski, Johann-Friedrich Feiden, and Carsten Rother.
\newblock Controlnet-xs: Rethinking the control of text-to-image diffusion
  models as feedback-control systems.
\newblock In \emph{European Conference on Computer Vision}, pages 343--362.
  Springer, 2024.

\bibitem[Zhang et~al.(2023)Zhang, Rao, and Agrawala]{zhang2023adding}
Lvmin Zhang, Anyi Rao, and Maneesh Agrawala.
\newblock Adding conditional control to text-to-image diffusion models.
\newblock In \emph{Proceedings of the IEEE/CVF International Conference on
  Computer Vision (ICCV'23)}, pages 3836--3847, 2023.

\bibitem[Zhao et~al.(2023)Zhao, Chen, Chen, Bao, Hao, Yuan, and
  Wong]{zhao2023uni}
Shihao Zhao, Dongdong Chen, Yen-Chun Chen, Jianmin Bao, Shaozhe Hao, Lu Yuan,
  and Kwan-Yee~K Wong.
\newblock Uni-controlnet: All-in-one control to text-to-image diffusion models.
\newblock \emph{Advances in Neural Information Processing Systems},
  36:\penalty0 11127--11150, 2023.

\bibitem[Zhou et~al.(2022)Zhou, Lei, Liu, Du, Huang, Zhao, Dai, Le, Laudon,
  et~al.]{zhou2022mixture}
Yanqi Zhou, Tao Lei, Hanxiao Liu, Nan Du, Yanping Huang, Vincent Zhao, Andrew~M
  Dai, Quoc~V Le, James Laudon, et~al.
\newblock Mixture-of-experts with expert choice routing.
\newblock \emph{Advances in Neural Information Processing Systems},
  35:\penalty0 7103--7114, 2022.

\end{thebibliography}
}

\clearpage
\appendix
\counterwithin{figure}{section}
\counterwithin{table}{section}
\counterwithin{equation}{section}
\section{More Details on Methods}
\subsection{Dynamic Gate with Loss-Free Balancing}
To ensure stable knowledge diversion across multiple tailors, we adopt a lightweight balancing strategy introduced in~\cite{wang2024auxiliary} that promotes uniform activation without relying on auxiliary loss terms. 
This mechanism mitigates over-reliance on a subset of tailors and encourages balanced utilization, thereby enhancing specialization and condition-specific diversity.

For a given condition with instruction $e_\text{txt}$, the activation scores $\alpha = [s_1, s_2, ..., s_{N_T}]$ are computed as described in Eq.~\eqref{eq:router}. Each tailor $T_i$ is assigned a learnable bias $b_i$, which is added to its activation score $s_i$ derived from the condition embedding. The dynamic gate selects the top-$K$ tailors based on the biased scores $s_i + b_i$:
\begin{equation}
    g_i =
    \begin{cases}
    s_i, & \text{if } s_i + b_i \in \text{Top-}K(\{s_j + b_j\}_{j=1}^{N_T}) \\
    0, & \text{otherwise}
    \end{cases}
\end{equation}
This mechanism adjusts tailor activation without affecting output magnitudes or introducing gradient interference.
The bias terms ${b_i}$ are updated after each batch based on usage statistics. Overused tailors have their biases reduced, while underused ones are increased to encourage activation. This adaptive process maintains load balance without affecting the training objective or model stability.

\subsection{More Details on Knowledge Diversion}
Algorithm~\ref{alg:algorithm} presents the pseudo-code for diverting condition-agnostic knowledge into reusable \textit{learngenes} and condition-specific knowledge into lightweight \textit{tailors} via structured SVD-based decomposition during multi-condition training.
\begin{algorithm}[h]
    \caption{Knowledge Diversion in Controllable Image Generation}
    \small
    \label{alg:algorithm}
    \textbf{Input}: ControlNet $\mathcal{F}$, training dataset $\mathcal{D} = \{(x^{(i)}, y^{(i)}, c^{(i)})\}_{i=1}^m$ with $N_{\text{basic}}$ basic conditions, number of epochs $N_{\text{ep}}$, batch size $B$, learning rate $\alpha$\\
    \textbf{Output}: Learngene $\mathcal{G}$
    \begin{algorithmic}[1]
        \STATE Randomly initialize the weight matrices $\mathcal{W}$ of $\mathcal{F}$, as well as the matrices $U_{\star}^{(l)}$, $\Sigma_{\star}^{(l)}$, and $V_{\star}^{(l)}$, and dynamic gate $G$
        \FOR{$ep = 1$ to $N_{\text{ep}}$}
            \FOR{each batch $\{(x_i, y_i)\}_{i=1}^B$}
                \STATE Update $\mathcal{W}$ of $f$ with $U_{\star}^{(l)}$, $\Sigma_{\star}^{(l)}$ and $V_{\star}^{(l)}$ under the rule of Eq.~\eqref{equ:svd}
                \STATE Compute routing weights $\alpha_i = G(e_{c_i})$ using Eq.~\eqref{eq:router}
                \STATE For each $x_i$, forward propagate $\hat{y}_i = \mathcal{F}(x_i, \alpha\cdot \theta)$
                \STATE Calculate $\mathcal{L}_{\text{batch}}=\frac{1}{B} \sum_{i=1}^B \mathcal{L}(\hat{y}_i, y_i)$ according to Eq.~\eqref{eq:loss}
                \STATE Backward propagate the loss $\mathcal{L}(\hat{y}_i, y_i)$ to compute the gradients with respect to $U_{\star}^{(l)}$, $\Sigma_{\star}^{(l)}$ and $V_{\star}^{(l)}$: $\nabla_{U} \mathcal{L}_{\text{batch}}, \nabla_{\Sigma} \mathcal{L}_{\text{batch}}$ and $\nabla_{V}\mathcal{L}_{\text{batch}}$
                \STATE Update the learngenes $U_{G,\star}^{(l)}$, $\Sigma_{G,\star}^{(l)}$ and $V_{G,\star}^{(l)}$:
                \\ \quad $U_{G,\star}^{(l)} := U_{G,\star}^{(l)} - \alpha \cdot \nabla_{U} \mathcal{L}_{\text{batch}}$, 
                \\ \quad $\Sigma_{G,\star}^{(l)} := \Sigma_{G,\star}^{(l)} - \alpha \cdot \nabla_{\Sigma} \mathcal{L}_{\text{batch}}$
                \\ \quad $V_{G,\star}^{(l)} := V_{G,\star}^{(l)} - \alpha \cdot \nabla_{V} \mathcal{L}_{\text{batch}}$
                \STATE Update the tailors $U_{T_i,\star}^{(l)}$, $\Sigma_{T_i,\star}^{(l)}$ and $V_{T_i,\star}^{(l)}$:
                \\ \quad $U_{T_i,\star}^{(l)} := U_{T_i,\star}^{(l)} - \alpha \cdot G(\nabla_{U} \mathcal{L}_{\text{batch}})$
                \\ \quad $\Sigma_{T_i,\star}^{(l)} := \Sigma_{T_i,\star}^{(l)} - \alpha \cdot G(\nabla_{\Sigma} \mathcal{L}_{\text{batch}})$
                \\ \quad $V_{T_i,\star}^{(l)} := V_{T_i,\star}^{(l)} - \alpha \cdot G(\nabla_{V} \mathcal{L}_{\text{batch}})$
            \ENDFOR
        \ENDFOR
    \end{algorithmic}
\end{algorithm}

\section{Training Details}
\subsection{Dataset}
\label{app:dataset}
\begin{figure*}[tb]
  \centering
  \includegraphics[width=\linewidth]{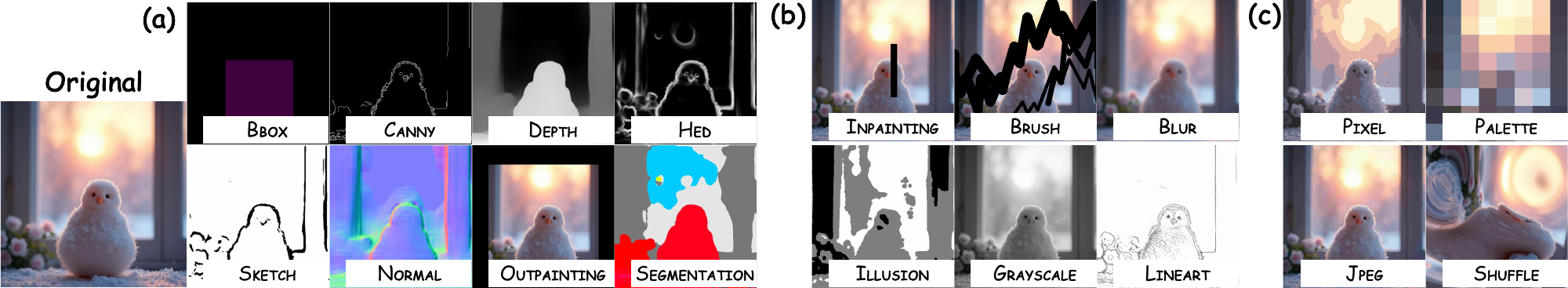}
  \vspace{-0.25in}
  \caption{\textbf{Conditions for Training and Evaluation.} (a) Eight basic conditions used for knowledge diversion and unified controllable generation. (b) Low-shift novel conditions for evaluating zero-shot generalization. (c) High-shift novel conditions for assessing few-shot adaptation.}
  \label{fig:app_dataset}
  \vspace{-0.1in}
\end{figure*}

To enable controllable generation and condition generalization, we construct a diverse set of conditions, as illustrated in Fig.~\ref{fig:app_dataset}. Basic conditions are used for learngene-driven knowledge diversion, while novel conditions are introduced to evaluate zero-shot and few-shot generalization.

\subsection{Hyper-parameters}
\label{app:hyper}
Table~\ref{tab:hyper_main} presents the basic settings, including learning rate, training steps and the number of learngene components $N_G$ and tailor components $N_T$ for DivControl diverting knowledge.

\begin{table}[tb]
    \centering
    \caption{Hyper-parameters for DivControl diverting knowledge on basic conditions.}
    \vspace{0.05in}
    \setlength{\tabcolsep}{17 mm}
        \begin{tabular}{@{}lr@{}}
        \toprule[1.5pt]
        \textbf{Training Settings} & \textbf{Configuration} \\
        \midrule[1.1pt]
        optimizer & AdamW \\
        scheduler & MultiStepLR \\
        decay step & 300K \\
        factor & 0.4 \\
        learning rate & 1.25e-5\\
        weight decay & 3e-2\\
        batch size & 16\\
        training steps & 450,000\\
        image size & 512$\times$512\\
        dropout & 0.1 \\
        $N_G$ & 512\\
        $N_T$ & 512\\
        REPA layer & 4\\
        $\lambda$ & 0.05 \\
        \bottomrule[1.5pt]
        \end{tabular}
    \label{tab:hyper_main}
\end{table}

\section{Additional Results}
\subsection{More Results on Zero-shot Generalization}
\begin{figure*}[tb]
  \centering
  \includegraphics[width=\linewidth]{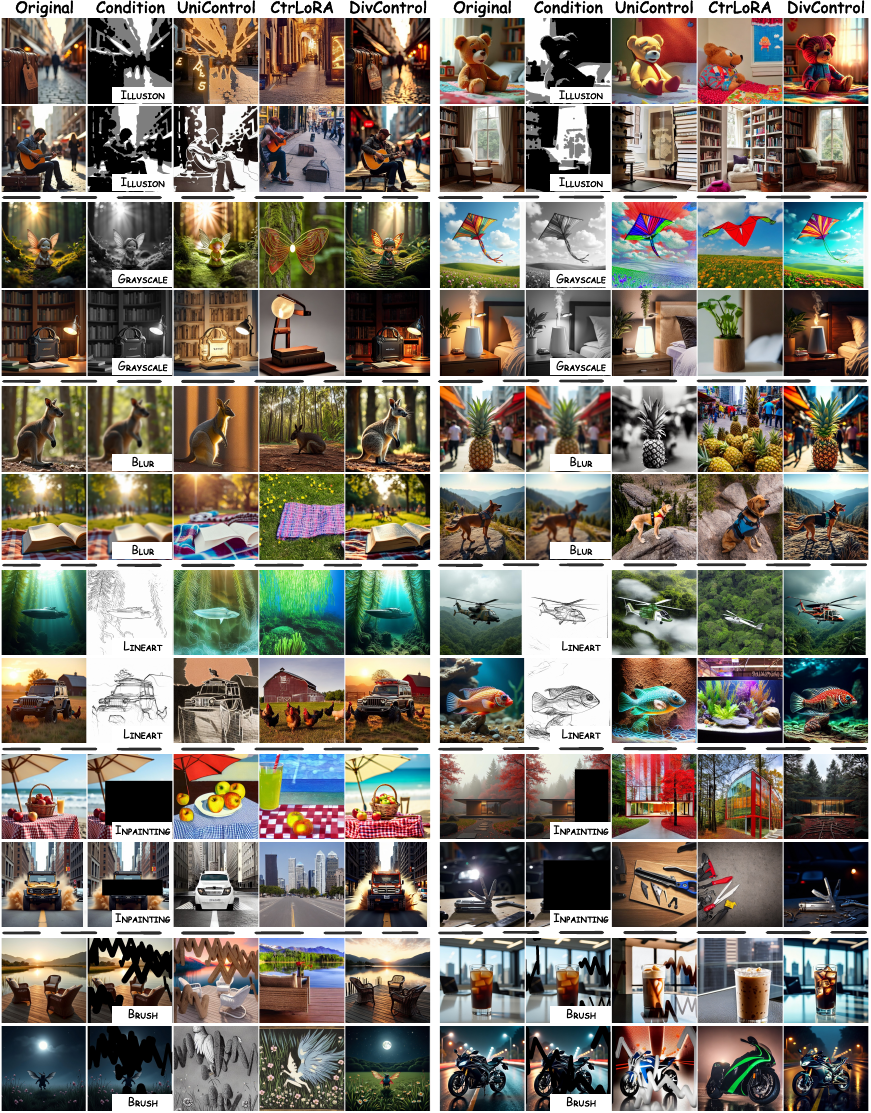}
  \vspace{-0.25in}
  \caption{\textbf{More Results on Zero-shot Generalization on Low-shift Novel Conditions.}}
  \label{fig:app_zero}
  \vspace{-0.1in}
\end{figure*}

To further validate the zero-shot generalization ability of DivControl, we provide additional qualitative and quantitative results under novel control conditions not seen during training. As shown in Fig.~\ref{fig:app_zero}, DivControl consistently maintains semantic alignment and visual fidelity without any adaptation, demonstrating its strong transferability and robustness under diverse unseen conditions.

\subsection{More Results on Multi-Conditional Controllable Generation}
\begin{figure*}[tb]
  \centering
  \includegraphics[width=\linewidth]{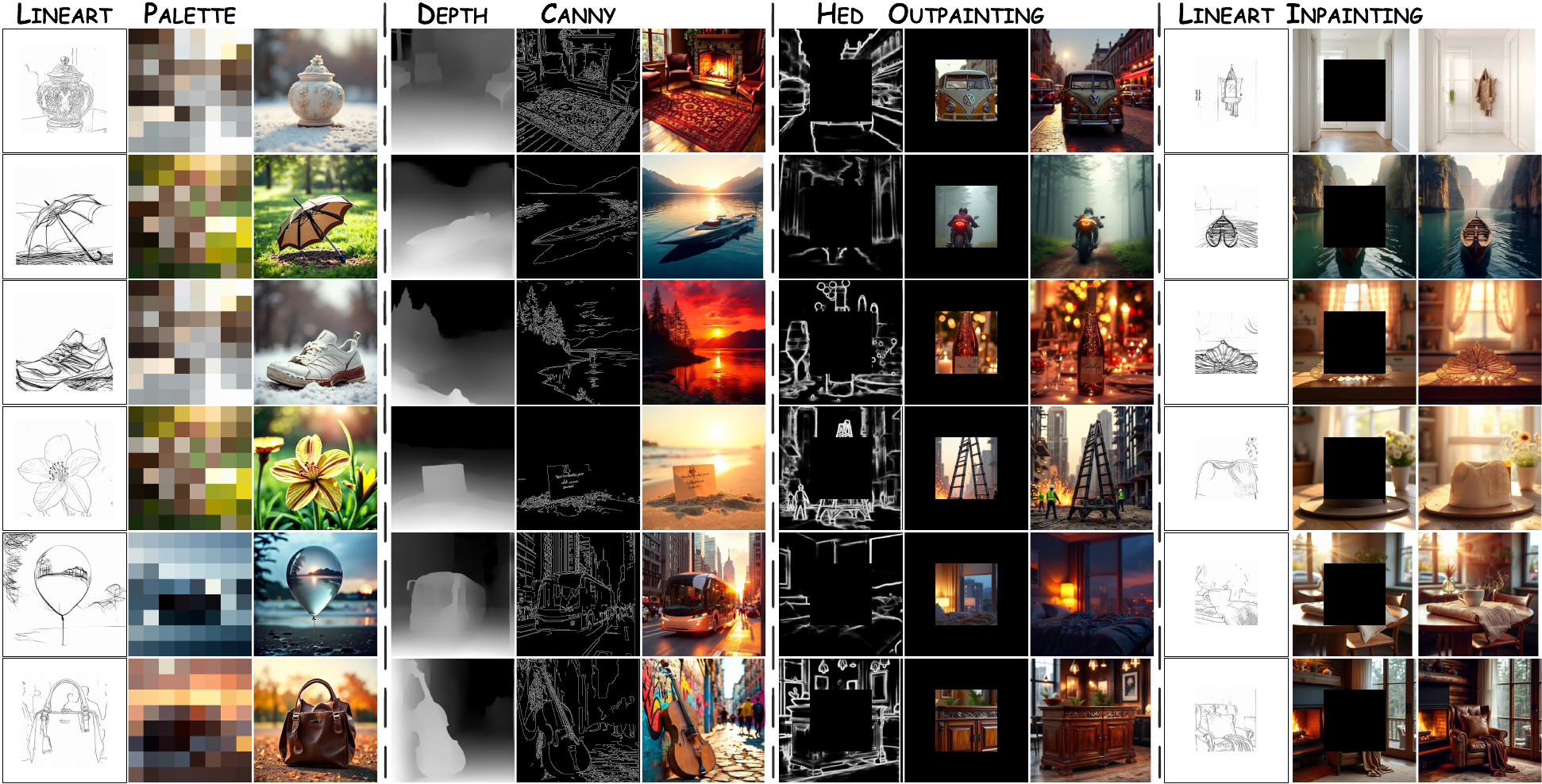}
  \vspace{-0.25in}
  \caption{\textbf{More Results on Multi-Conditional Controllable Image Generation.}}
  \label{fig:app_multi}
  \vspace{-0.1in}
\end{figure*}
We present additional results on multi-conditional controllable generation, showcasing DivControl's ability to integrate multiple control conditions. We evaluate various condition combinations, including \textsc{Lineart}, \textsc{Palette}, and \textsc{Outpainting}, demonstrating DivControl's seamless synthesis of these conditions into coherent outputs. As shown in Fig.~\ref{fig:app_multi}, DivControl consistently generates high-quality, semantically aligned images, underscoring its flexibility and efficient feature reuse.

\end{document}